\documentclass[journal]{elsarticle}

\usepackage{lineno,hyperref}
\usepackage{amsmath}
\usepackage{algorithm}
\usepackage{algorithmic}
\usepackage{amssymb}
\modulolinenumbers[5]

\journal{Journal of \LaTeX\ Templates}

\begin{document}

\begin{frontmatter}

\title{Generating Random Parameters in Feedforward Neural Networks with Random Hidden Nodes: Drawbacks of the Standard Method and How to Improve It
}
\tnotetext[mytitlenote]{Supported by Grant 2017/27/B/ST6/01804 from the National Science Centre, Poland.}

\author{Grzegorz Dudek}
\address{Electrical Engineering Department, Czestochowa University of Technology, Czestochowa, Poland, dudek@el.pcz.czest.pl}




\begin{abstract}

The standard method of generating random weights and biases in feedforward neural networks with random hidden nodes, selects them both from the uniform distribution over the same fixed interval. In this work, we show the drawbacks of this approach and propose a new method of generating random parameters. This method ensures the most nonlinear fragments of sigmoids, which are most useful in modeling target function nonlinearity, are kept in the input hypercube. In addition, we show how to generate activation functions with uniformly distributed slope angles. 
\end{abstract}


\end{frontmatter}


\section{Introduction}
Single-hidden-layer feedforward neural networks with random hidden nodes (FNNRHN) have become popular in recent years due to their fast learning speed, good generalization performance and ease of implementation. Additionally, these networks do not use a gradient descent method for learning, which is time consuming and sensitive to local minima of the error function (which is nonconvex in this case). In randomized learning, weights and biases of the hidden nodes are selected at random from any interval $[-u,u]$, and stay fixed. The optimization problem becomes convex and the output weights can be learned using a simple, scalable standard linear least-squares method \cite{Pri15}. The resulting FNN has a universal approximation capability when the random parameters are selected from a symmetric interval according to any continuous sampling distribution \cite{Hau99}. But how to select this interval and which distribution to use are open questions, and considered to be the most important research gaps in randomized learning \cite{Zha16s,Cao18}. 

Typically, the hidden node weights and biases are both selected from a uniform distribution over the fixed interval, $[-1,1]$,  without scientific justification, regardless of the data, problem to be solved, and activation function type \cite{Scar16}. Some authors optimize the interval looking for $u$ to ensure the best model performance \cite{wang17,Li17,Zha16,Cao16}. Recently developed methods \cite{dud19,dud19a} propose more sophisticated approaches for generating random parameters, where the distribution of the activation functions in space is analyzed and their parameters are adjusted randomly to the data.

In this work we show the drawbacks of a standard method of random parameters generation and propose its modification. We treat the weights and biases separately due to their different functions. The biases are generated on the basis of the weights and points selected randomly from the input space. The resulting sigmoids have their nonlinear fragments, which are most useful for modeling the target function (TF) fluctuations, inside the input hypercube. Moreover, we show how to generate the weights to produce sigmoids with the slope angles distributed uniformly.            

\section{Generating sigmoids inside the input hypercube}

Let us consider an approximation problem of a single-variable function of the form:

\begin{equation}
g(x) = \sin\left(20\cdot\exp x \right)\cdot x^2
\label{eqTF1}
\end{equation}
To learn FNNRHN we create a training set $\Phi$ containing $N = 5000 $ points $ (x_l, y_l) $, where $ x_l \sim U(0,1) $ and $ y_l $  are calculated from \eqref{eqTF1} and then distorted by adding noise $ \xi \sim U(-0.2, 0.2) $. A test set of the same size is created in the same manner but without noise. The output is normalized in the range $ [-1, 1] $.

\begin{figure}
	\centering
	\includegraphics[width=0.49\textwidth]{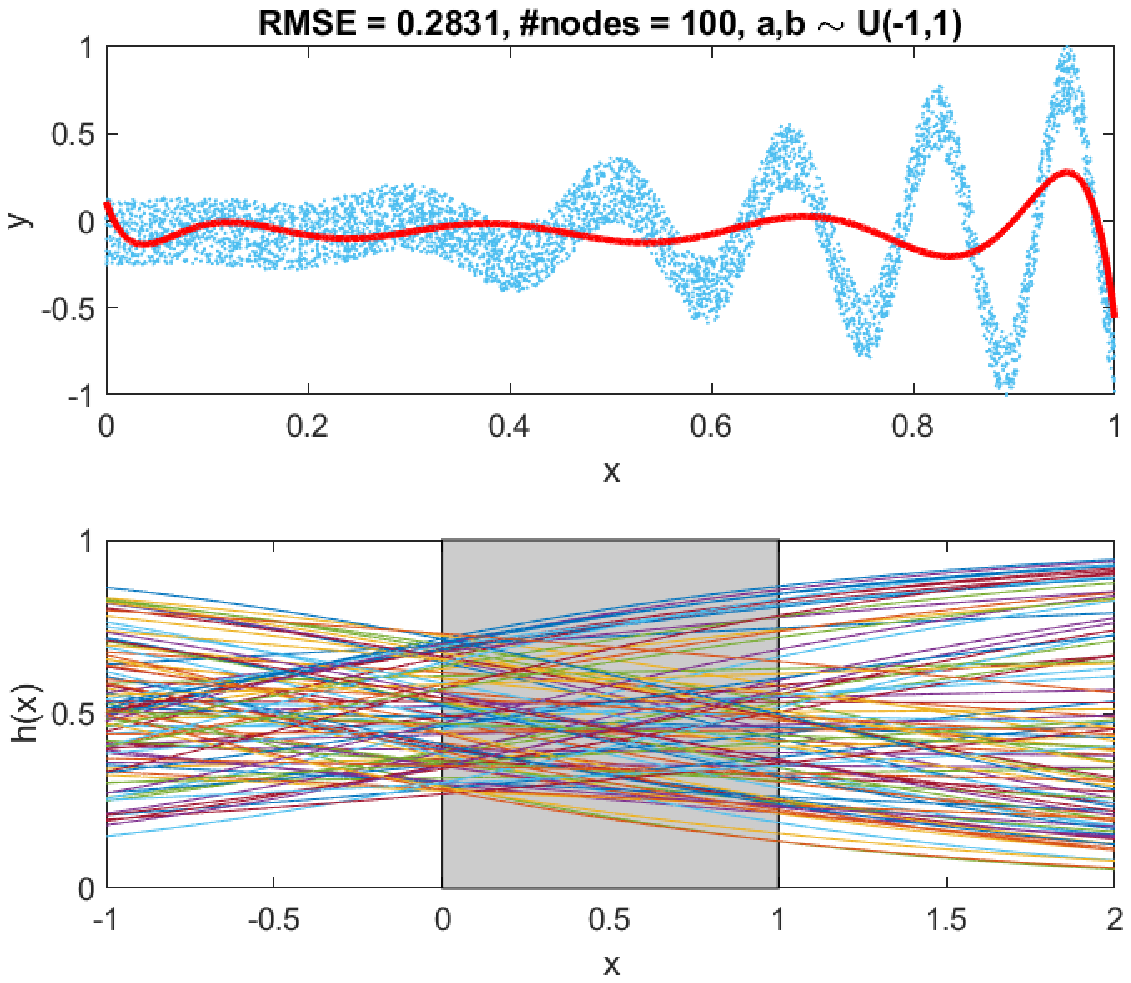}
	\includegraphics[width=0.49\textwidth]{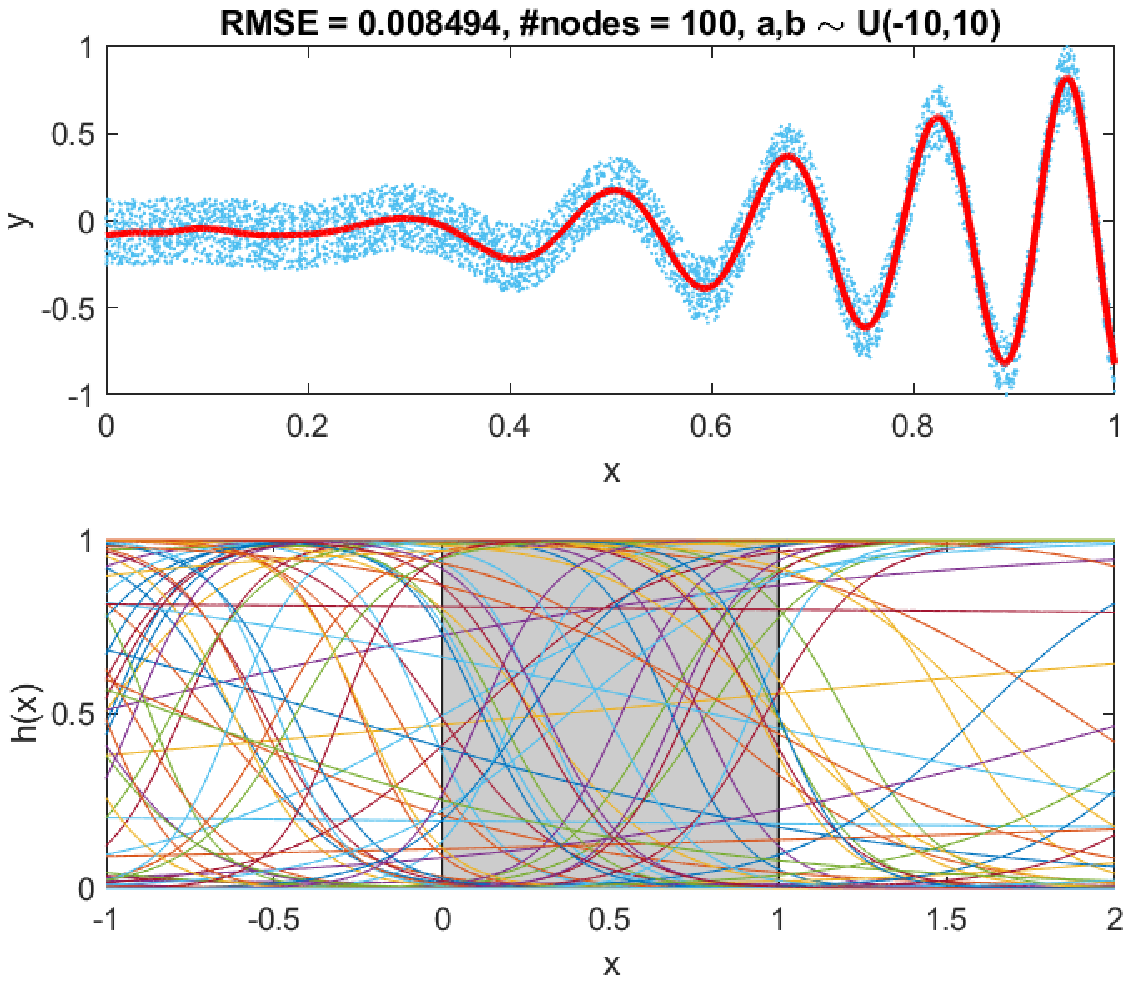}
	\caption{TF \eqref{eqTF1} fitting: fitted curves and the sigmoids constructing them for $a,b \sim U(-1,1)$ (left panel) and for $a,b \sim U(-10,10)$ (right panel).} 
	\label{fig0}
\end{figure}

Fig. \ref{fig0} shows the results of fitting when using FNNRHN with $100$ sigmoid hidden nodes where the weights and biases are randomly selected from $U(-1,1)$ and $U(-10, 10)$. The bottom charts show the hidden node sigmoids whose linear combination forms the function fitting data. This fitted function is shown as a solid line in the upper charts. As you can see from the figure, for $a,b \in [-1,1]$ the sigmoids are flat and their distribution in the input interval $[0,1]$ (shown as a grey field) does not correspond to the TF fluctuations. This results in a very weak fit. When $a,b \in [-10,10]$ the sigmoids are steeper but many of them have their steepest fragments, which are around their inflection points, outside of the input interval. The saturated fragments of these sigmoids, which are in the input interval, are useless for modeling nonlinear TFs. So, many of the $100$ sigmoids are wasted. From this simple example it can be concluded that to get a parsimonious flexible FNNRHN model, the sigmoids should be steep enough and their steepest fragments, around the inflection points, should be inside the input interval.     

Let us analyze how the inflection points are distributed in space when the weights and biases are selected from a uniform distribution over the interval $[-u, u]$. The sigmoid value at its inflection point $\chi$ is $0.5$, thus:

\begin{equation}
\frac{1}{1 + \exp(-(a\cdot \chi + b))} = 0.5
\label{eq2}
\end{equation}

From this equation we obtain:

\begin{equation}
\chi = -\frac{b}{a}
\label{eq3}
\end{equation}

The distribution of the inflection point is a distribution of the ratio of two independent random variables having both the uniform distribution, $a,b \sim U(-u, u)$. In such a case, the probability density function (PDF) of $\chi$ is of the form:

\begin{equation}
\begin{split}
f(\chi) & =  \int_{-\infty}^{\infty} |a|f_A(a)f_B(a\chi)da\\
&=\left\lbrace 
\begin{array}{lll}
\displaystyle\int_{-u}^{u} |a|f_A(a)f_B(a\chi)da & \mathrm{for} & |\chi|<1 \\
\displaystyle\int_{-\frac{u}{|\chi|}}^{\frac{u}{|\chi|}} |a|f_A(a)f_B(a\chi)da & \mathrm{for} & |\chi|\geq 1
\end{array}
\right.\\
&=\left\lbrace 
\begin{array}{lll}
\displaystyle\frac{1}{4} & \mathrm{for} & |\chi|<1\\
\displaystyle\frac{1}{4|\chi|^2} & \mathrm{for} & |\chi|\geq 1
\end{array}
\right.
\end{split}
\label{eq4}
\end{equation}   
where $f_A$ and $f_B$ are the PDFs of weights and biases, respectively. 

\begin{figure}
	\centering
	\includegraphics[width=0.49\textwidth]{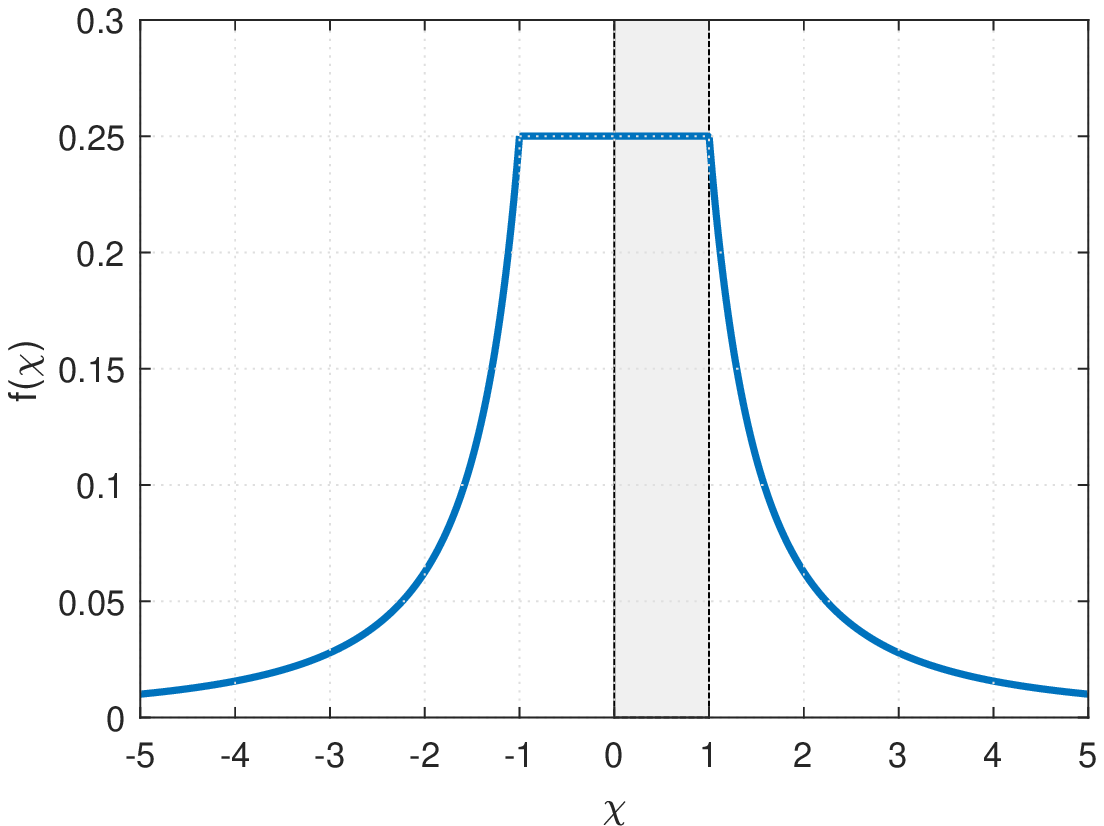}
	\includegraphics[width=0.49\textwidth]{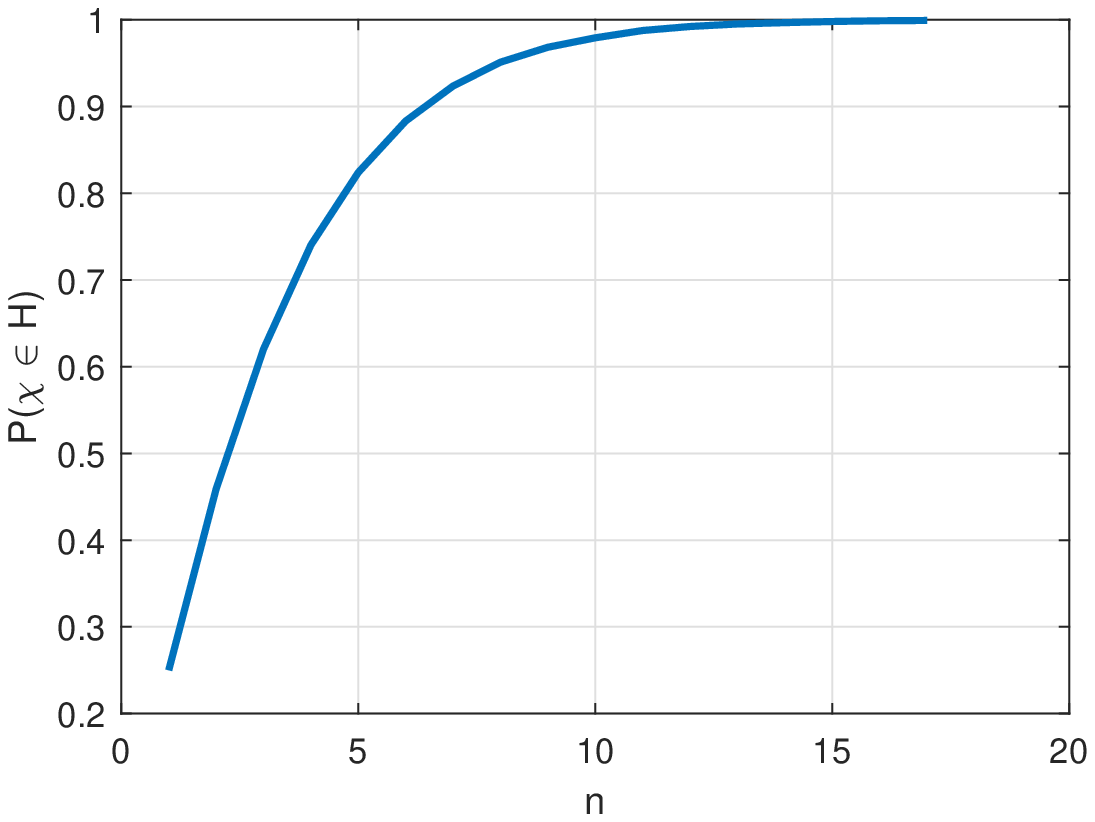}
	\caption{PDF of $\chi$ when $a,b \sim U(-u,u)$ (left panel) and probability that $\chi$ belongs to $H=[0,1]^n$ depending on $n$ (right panel).} 
	\label{fig1}
\end{figure}

The left panel of Fig. \ref{fig1} shows the PDF of $\chi$. The same PDF can be obtained when $a \sim U(-u, u)$ and $b \sim U(0, u)$ (case sometimes found in the literature). As you can see from Fig. \ref{fig1}, the probability that the inflection point is inside the input interval (shown as a grey field) is $0.25$. This means that most sigmoids have their steepest fragments, which are most useful for modeling TF fluctuations, outside of this interval. For the multivariable case, when we consider $n$-dimensional sigmoids, the situation improves -- see the right panel of Fig. \ref{fig1}. For $n=2$ almost $46\%$ of sigmoids have their inflection points in the input square. This percentage increases to more than $90\%$ for $n \geq 7$.   

To obtain an $n$-dimensional sigmoid with one of its inflection points $\chi$ inside the input hypercube $ H = [x_{1,\min}, x_{1,\max}]\times ... \times[x_{n,\min}, x_{n,\max}] $, first, we generate weights $\mathbf{a}=[a_1, a_2,..., a_n]^T \subset \mathbb{R}^n$. Then we set the sigmoid in such a way that $\chi$ is at some point $\mathbf{x}^*$ from $H$. Thus: 

\begin{equation}
h(\mathbf{x}^*) = \frac{1}{1 + \exp\left(-\left(\mathbf{a}^T\mathbf{x}^* + b\right)\right)}=0.5
\label{eqSigM}
\end{equation}

From this equation we obtain:

\begin{equation}
b = -\mathbf{a}^T\mathbf{x}^*
\label{eqDer5a}
\end{equation}
  
Point $\mathbf{x}^*=[x_1^*, ..., x_n^*]$ can be selected as follows:
\begin{itemize}
	\item this can be some point randomly selected from $H$: $ x_j^* \sim U(x_{j,\min}, x_{j,\max})$, $j=1,2,...,n$. This method is suitable when the input points are evenly distributed in the hypercube $ H $.
	\item this can be some randomly selected training point: $\mathbf{x}^* = \mathbf{x}_\xi \in \Phi $, where $\xi \sim U\{1, ..., N\}$. This methods distributes the sigmoids according to the data density, avoiding empty regions.
	\item this can be a prototype of the training point cluster: $\mathbf{x}^* = \mathbf{p}_i $, where $ \mathbf{p}_i $ is a~prototype of the $ i $-th cluster of $ \mathbf{x} \in \Phi $. This method groups the training points into $ m =$\#nodes clusters. For each sigmoid a different prototype is taken as $\mathbf{x}^*$.   
\end{itemize}

\section{Generating sigmoids with uniformly distributed slope angles}

\begin{figure}
	\centering
	\includegraphics[width=0.49\textwidth]{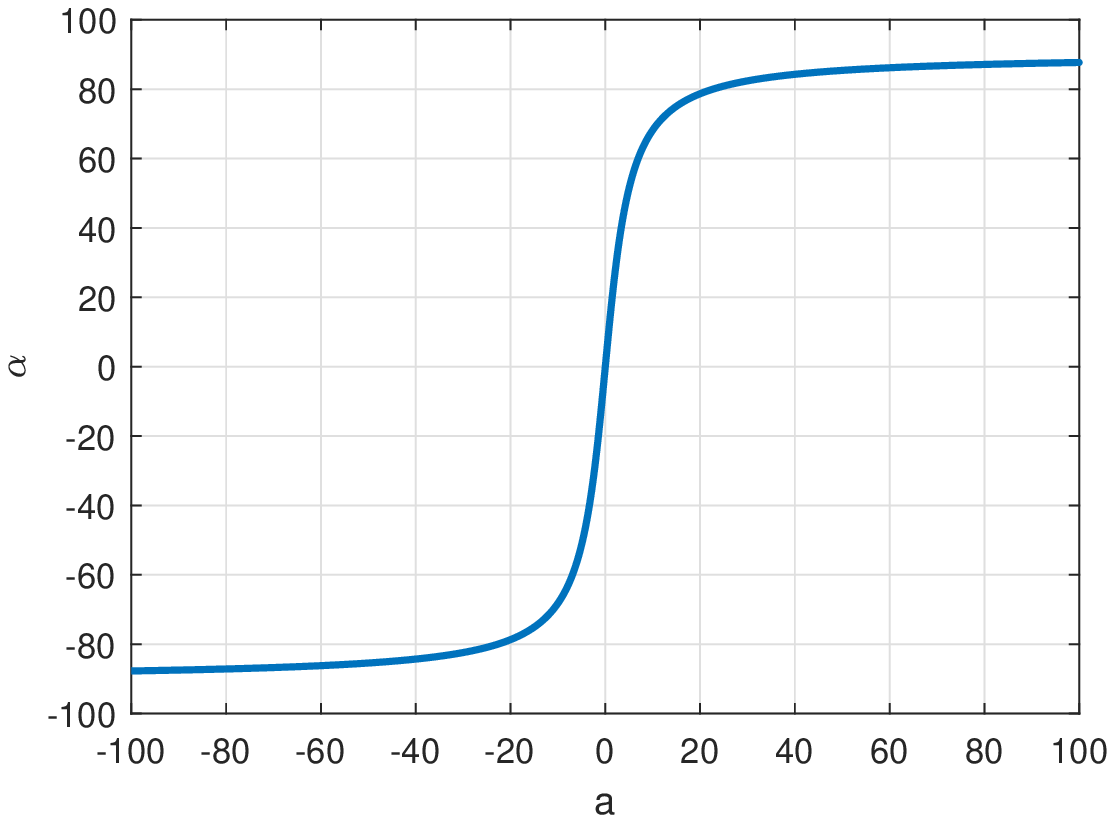}
	\includegraphics[width=0.49\textwidth]{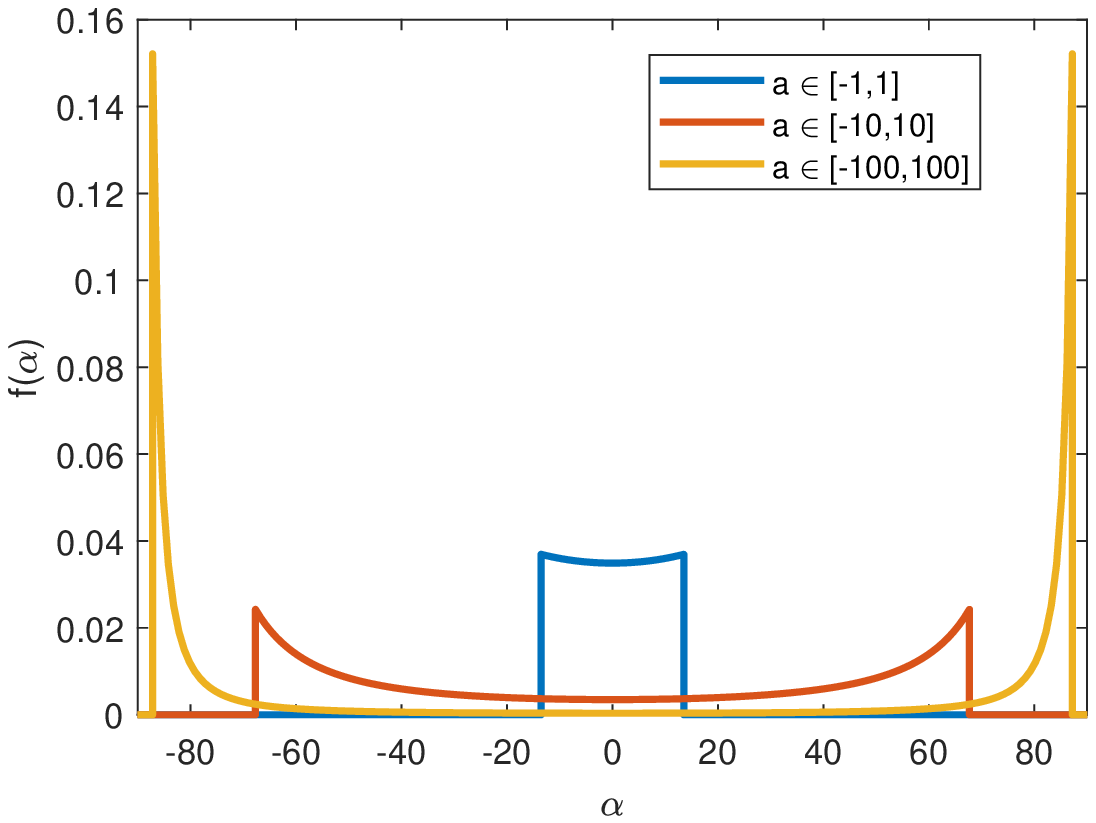}
	\caption{Relationship between $a$ and $\alpha$ (left panel) and PDF of $\alpha$ for different intervals for $a$ (right panel).} 
	\label{fig5}
\end{figure}

It should be noted that weight $a$ translates nonlinearly into the slope angle of a sigmoid. Let us analyze sigmoid $S$ which has its inflection point $\chi$ at $x=0$. In such a case $b=0$. A derivative of $S$ at $x=0$ is equal to the tangent of its slope angle $\alpha$ at $\chi$:
\begin{equation}
\begin{split}
\tan\alpha & = ah(x)\left(1-h(x)\right) \\
& = a\frac{1}{1 + \exp(-(a\cdot 0 + 0))} \left(1-\frac{1}{1 + \exp(-(a\cdot 0 + 0))}\right) 
\end{split}
\label{eq5}
\end{equation}
From this equation we obtain the relationship between the weight and the slope angle:
\begin{equation}
\alpha = \arctan \displaystyle\frac{a}{4} 
\label{eq6}
\end{equation}
This relationship is depicted in Fig. \ref{fig5} as well as the PDF of $\alpha$ when weights $a$ are generated from different intervals. Note that the relationship between $a$ and $\alpha$ is highly nonlinear. Interval  $[-1, 1]$ for $a$ corresponds to the interval $[-14^\circ, 14^\circ]$ for $\alpha$, so only flat sigmoids are obtainable in such a case. For $a \in [-10, 10]$ we obtain  $\alpha \in [-68.2^\circ, 68.2^\circ]$, and for $a \in [-100, 100]$ we obtain  $\alpha \in [-87.7^\circ, 87.7^\circ]$. For narrow intervals for $a$, such as $[-1, 1]$, the distribution of $\alpha$ is similar to a uniform one. When the interval for $a$ is extended, the shape of PDF of $\alpha$ changes -- larger angles, near the bounds, are more probable than smaller ones. When $a \in [-100, 100]$ more than $77\%$ of sigmoids are inclined at an angle greater than $80^\circ$, so they are very steep. In such a case, there is a real threat of overfitting.   

To generate sigmoids with uniformly distributed slope angles, first we generate $|\alpha| \sim U(\alpha_{\min}, \alpha_{\max})$ individually for them, where $\alpha_{\min} \in (0^\circ, 90^\circ)$ and $\alpha_{\max} \in (\alpha_{\min}, 90^\circ)$. The border angles, $\alpha_{\min}$ and $\alpha_{\max}$, can both be adjusted to the problem being solved. For highly nonlinear TFs, with strong fluctuations, only $\alpha_{\min}$ can be adjusted, keeping $\alpha_{\max}=90^\circ$. Having the angles, we calculate the weights from \eqref{eq6}:
 
 \begin{equation}
 a=4 \tan \alpha 
 \label{eq6a}
 \end{equation}

For the multivariable case, we generate all $n$ weights in this way, independently for each of $m$ sigmoids. This ensures random slopes (between $\alpha_{\min}$ and $\alpha_{\max}$) for the multidimensional sigmoids in each of $n$ directions.   

The proposed method of generating random parameters of the hidden neurons is summarized in Algorithm 1. In this algorithm weights $a$ can be generated randomly from $U(-u, u)$ or optionally, to ensure uniform distribution of the sigmoid slope angles, they can be determined based on the slope angles generated randomly from $U(\alpha_{\min},\alpha_{\max})$. The bounds: $u, \alpha_{\min}$ and $\alpha_{\max}$ should be selected in cross-validation. 

\begin{algorithm}[H]
	\caption{Generating Random Parameters of FNNRHN}
	\label{alg1}
	\begin{algorithmic}
		\vspace{4mm}
		\STATE {\bfseries Input:}\\ 
		\vspace{4mm}
		Number of hidden nodes $m$\\
		Number of inputs $n$\\
		Bounds for weights, $u \in \mathbb{R}$, or optionally bounds for slope angles, \\
		$\alpha_{\min} \in (0^\circ, 90^\circ)$ and $\alpha_{\max} \in (\alpha_{\min}, 90^\circ)$ \\ 
		Set of $m$ points $\mathbf{x}^* \in H$: $\{\mathbf{x}^*_1, ..., \mathbf{x}^*_m\}$ \\
		\vspace{4mm}
		\STATE {\bfseries Output:}\\ 
		\vspace{4mm}
		\hspace{4mm} Weights $ \mathbf{A} = \left[
		\begin{array}{ccc}
		a_{1,1} & \ldots & a_{m,1} \\
		\vdots & \ddots & \vdots \\
		a_{1,n} & \ldots & a_{m,n}
		\end{array}
		\right]	$  \\
		\hspace{4mm} Biases $ \mathbf{b} = [b_1, \ldots, b_m] $ \\    
		\vspace{4mm}
		\STATE {\bfseries Procedure:}\\
		\vspace{4mm}
		\FOR{$i=1$ {\bfseries to} $m$}
		\FOR{$j=1$ {\bfseries to} $n$}
		\STATE Choose randomly $a_{i,j} \sim U(-u,u)$ \\
		\STATE or optionally choose randomly $\alpha_{i,j} \sim U(\alpha_{\min},\alpha_{\max}) $ and calculate
		\begin{equation*}
		\begin{split}
		a_{i,j} = (-1)^q \cdot 4\tan \alpha_{i,j},\ \textrm{where}\ q \sim U\{0,1\} 		
		\end{split}
		\end{equation*}

		\ENDFOR
		
		\STATE Calculate
		\begin{equation*}
		b_i = -\mathbf{a}^T_i\mathbf{x}^*_i
		\end{equation*}
		
		\ENDFOR
	\end{algorithmic}
\end{algorithm}

\section{Simulation study}

\begin{figure}
	\centering
	\includegraphics[width=0.49\textwidth]{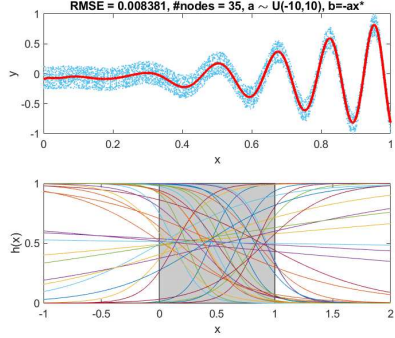}
	\caption{TF \eqref{eqTF1} fitting: fitted curve and the sigmoids constructing it for the proposed method.} 
	\label{fig6}
\end{figure}

The results of TF \eqref{eqTF1} fitting when using the proposed method is shown in Fig. \ref{fig6}. In this case the weights were selected from $U(-10,10)$ and biases were determined according to \eqref{eq6}. As you can see from this figure, all sigmoids have their inflection points inside $H$. The number of hidden nodes to achieve $RMSE=0.0084$ is 35. To obtain a similar level of error we need over 60 nodes when using the standard method for generating the parameters.

The following experiments concern multivariable function fitting. TF in this case is defined as:

\begin{equation}
g(\mathbf{x}) = \sum_{j=1}^{n}\sin\left(20\cdot\exp x_j\right)\cdot x_j^2
\label{eqTF2}
\end{equation}

The training set contains $N$ points $ (\mathbf{x}_l, y_l) $, where $ x_{l,j} \sim U(0,1) $ and $ y_l $  are calculated from \eqref{eqTF2}, then normalized in the range $ [-1, 1]$ and distorted by adding noise $ \xi \sim U(-0.2, 0.2) $. A test set of the same size is created in the same manner but without noise.
   
The experiments were carried out for $n=2$ ($N=5000$), $n=5$ ($N=20000$) and $n=10$ ($N=50000$), using:

\begin{itemize}
	\item SM -- the standard method of generating both weights and biases from $U(-u,u)$,
	\item PMu -- the proposed method of generating weights from $U(-u,u)$ and biases according to \eqref{eq6},
	\item PM$\alpha$ -- the proposed method of generating slope angles from $U(\alpha_{\min}, 90^\circ)$, then calculating weights from \eqref{eq6a}, and biases from \eqref{eq6}.  
\end{itemize}

\begin{figure}
	\centering
	\includegraphics[width=0.32\textwidth]{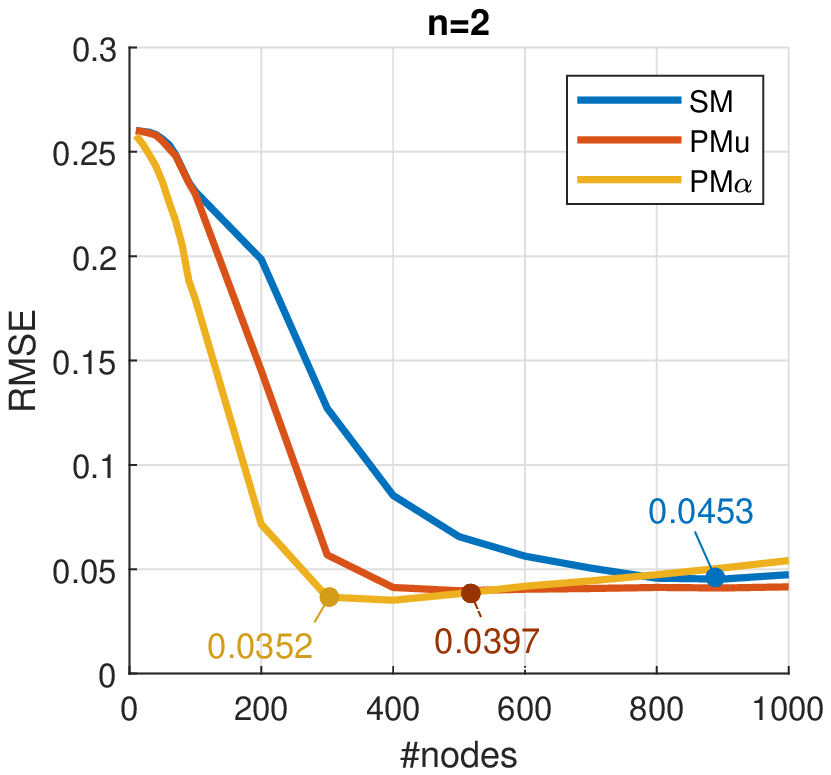}
	\includegraphics[width=0.32\textwidth]{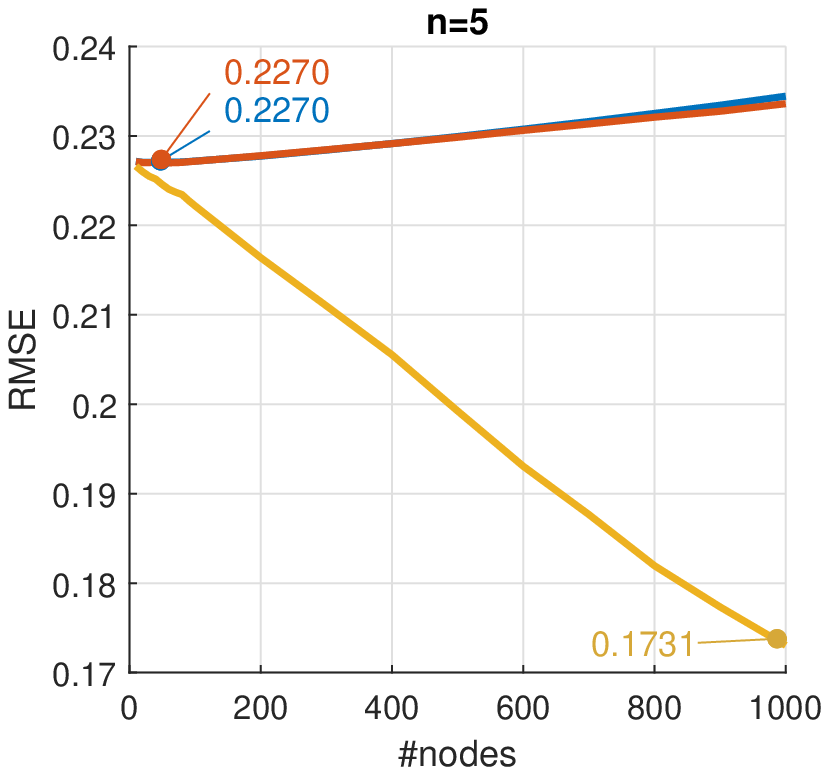}
	\includegraphics[width=0.32\textwidth]{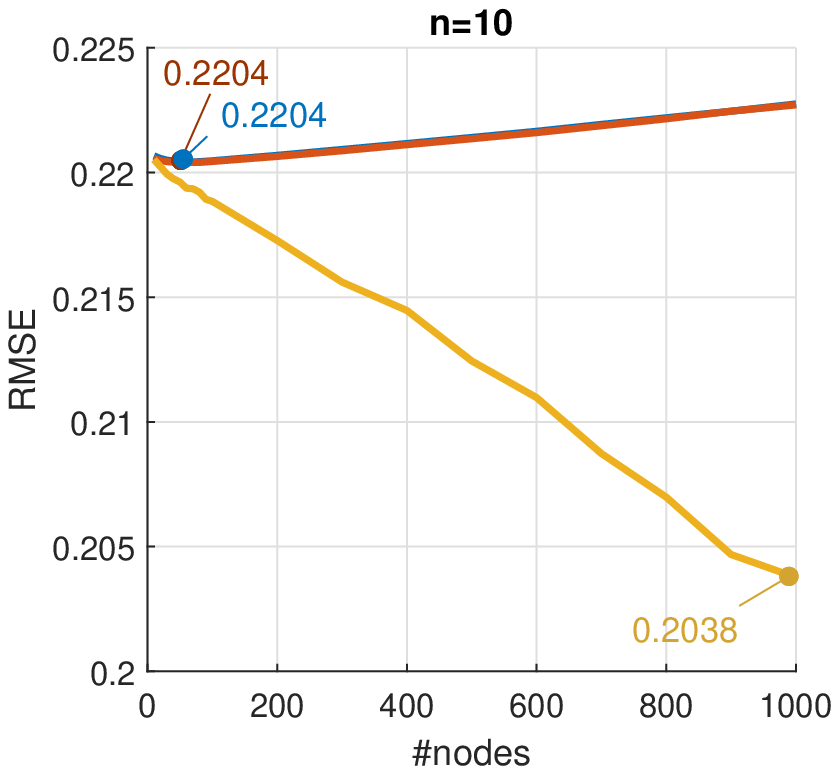}
	\caption{RMSE depending on the number of nodes.} 
	\label{fig7}
\end{figure}

Fig. \ref{fig7} shows the mean test errors over 100 trials for different node numbers. For each node number the optimal value of $u$ or $\alpha_{\min}$ was selected from $u \in \{1,2,...,10,20,50,100\}$ and $\alpha_{\min} \in \{0^\circ,10^\circ,...,80^\circ\}$, respectively. As you can see from Fig. \ref{fig7}, PM$\alpha$ in all cases leads to the best results. For $n=2$ it needs less nodes to get a lower error (0.0352) than PMu and SM. Interestingly, for higher dimensions, using too many nodes leads to an increase in the error for SM and PMu. This can be related to the overfitting caused by the steep nodes generated by the standard method. In the same time, for PM$\alpha$, where the node slope angles are distributed uniformly, an decrease in the error is observed. This issue needs to be explored in detail on a larger number of datasets.
      
\section{Conclusion}

A drawback of the standard method of generating random hidden nodes in FNNs is that many sigmoids have their most nonlinear fragments outside of the input hypercube, especially for low-dimensional cases. So, they cannot be used for modeling the target function fluctuations. In this work, we propose a method of generating random parameters which ensures that all the sigmoids have their steepest fragments inside the input hypercube. In addition, we show how to determine the weights to ensure the sigmoids have  uniformly distributed slope angles. This prevents overfitting which can happen when weights are generated in a standard way, especially for highly nonlinear target functions.


\bibliography{mybibfile}

\end{document}